\documentclass[runningheads]{llncs}
\usepackage{graphicx}
\usepackage{algorithm}
\usepackage{algpseudocode}
\usepackage{amssymb}
\usepackage{amsmath}
\newcommand{\argmin}{\operatornamewithlimits{arg\ min}}
\usepackage{eurosym}

\begin{document}

\title{Soccer Team Vectors}
%
%
\author{Robert M\"uller, Stefan Langer, Fabian Ritz, Christoph Roch, \\ Steffen Illium and Claudia Linnhoff-Popien}
\authorrunning{R. M\"uller et al.}
%
\institute{Mobile and Distributed Systems Group\\LMU Munich\\
\email{\{robert.mueller,stefan.langer,fabian.ritz, christoph.roch,steffen.illium,linnhoff\}@ifi.lmu.de}
}
\maketitle              
\begin{abstract}
In this work we present STEVE - \textbf{S}occer \textbf{TE}am \textbf{VE}ctors, a principled approach for learning real valued vectors for soccer teams where similar teams are close to each other in the resulting vector space. \textit{STEVE} only relies on freely available information about the matches teams played in the past. These vectors can serve as input to various machine learning tasks. Evaluating on the task of team market value estimation, \textit{STEVE} outperforms all its competitors. Moreover, we use \textit{STEVE} for similarity search and to rank soccer teams.

\keywords{Sports Analytics \and Representation Learning \and Self-Supervised Learning \and Soccer \and Football}
\end{abstract}
\section{Introduction}
The field of soccer analytics suffers from poor availability of free and affordable data. While Northern American sports have already been the subject of data analytics for a long time, soccer analytics has only started to gain traction in the recent years. \\
Feature vectors usually serve as an input to machine learning models. They provide a numeric description of an objects characteristics. However, in the case of soccer analytics these features are hard to obtain. For example, collecting non-trivial features for a soccer player or a team involves buying data from a sports analysis company which employs experts to gather data.\\
In this work we propose STEVE - \textbf{S}occer \textbf{TE}am \textbf{VE}ctors, a method to automatically learn feature vectors of soccer teams. 
\textit{STEVE} is designed to only use freely available match results from different soccer leagues and competitions. Thus, we alleviate the problem of poor data availability in soccer analytics. Automatically extracted feature vectors are usually referred to as representations in the literature. These representations can conveniently serve as input to various machine learning tasks like classification, clustering and regression.
In the resulting vector space, similar teams are close to each other. We base the notion of similarity between soccer teams on four solid assumptions (Section \ref{sec:main}). The most important one is that two teams are similar if they often win against the same opponents. Hence, \textit{STEVE} can be used to find similar teams by computing the distance between representations and to rank a self chosen list of teams according to their strengths.\\

\noindent This paper is organized as follows: In Section \ref{sec:related_work} we review related work. It consists of an in depth discussion of the process of learning real valued vectors for elements in a set and its applications. We also briefly review a recent approach to team ranking. 
In Section \ref{sec:main} we introduce \textit{STEVE}, our approach to learn meaningful representations for soccer teams. After giving an overview of the underlying idea, we introduce the problem with more rigor and conclude the section by formulating an algorithm for the task.
In Section \ref{sec:experiments} we conduct various experiments to evaluate the approach. Finally, Section \ref{sec:conculsion} closes out with conclusions and outlines future work.

\section{Related Work}
\label{sec:related_work}
Learning real valued vectors for elements in a set has been been of particular interest in the field of natural language processing.
Elements are usually words or sentences and their representation is computed in such a way, that they entail their meaning.
Modern approaches typically learn a distributed representation for words \cite{Bengio2003} based on the distributional hypothesis \cite{Harris1954}, which states that words with similar meanings often occur in the same contexts.\\
Mikolov et al. \cite{Mikolov2013b,Mikolov2013a} introduced \textit{word2vec}, a neural language model which uses the skip-gram architecture to train word representations. Given a center word \textit{word2vec} by iteratively maximizes the probability of observing the surrounding window of context words. The resulting representations can be used to measure semantic similarity between words. According to \textit{word2vec} the most similar word to \textit{soccer} is \textit{football}. Moreover, vector arithmetic can be used to compute analogies. Although having recently been put in question \cite{Kalidindi2019,Allen2019}, a very famous example is the following: \textit{king - man + woman = queen}.
The concept has since then been extended to graph structured data to learn a representation for each node in a graph. Perozzi et al. \cite{Perozzi2014} and Dong et al. \cite{Dong2017} treat random walks as the equivalent of sentences. This is based on the assumption that these walks can be interpreted as sampling from a language graph. The resulting sentences are fed to \textit{word2vec}. Building upon graph based representation learning approaches, \textit{LinNet} \cite{Pelechrinis2018} builds a weighted directed match-up network where nodes represent lineups from NBA basketball teams. An edge from node $i$ to node $j$ is inserted if lineup $i$ outperformed lineup $j$. The edge weight is set to the performance margin of the corresponding match-up. Lineup representations are computed by deploying \textit{node2vec} \cite{Grover2016} on the resulting network. Afterwards, a logistic regression model based on the previously computed lineup representations is learned to model the probability of  lineup $\lambda_i$ outperforming lineup
$\lambda_i$.\\
More recently, the aforementioned findings have also been applied to sports analytics. (batter$\vert$pitcher)2vec \cite{Alcorn2016} computes representations of Major League Baseball players through a supervised learning task that predicts the outcome of an at-bat given the context of a specific batter and pitcher.
The resulting representations are used to cluster pitchers who rely on pitches with dramatic movement and predict future at-bat outcomes. Further, by performing simple arithmetic in the learned vector space they identify opposite-handed doppelgangers.\\
Le et. al \cite{Le2017} introduce a data-driven ghosting model based on tracking data of a season from a professional soccer league to generate the defensive motion patterns of a \textit{league average} team. To fine-tune the \textit{league average} model to account for a team's structural and stylistic elements, each team is associated with a team identity vector.\\
Our approach aims to learn representations for soccer teams and is thus closely related to the presented approaches. As we use the representations to rank teams, we briefly review related work on the topic.\\
Neumann et al. \cite{Neumann2018} propose an alternative to classical ELO and Pi rating based team ranking approaches \cite{Hvattum2010,Constantinou2013}. 
A graph based on the match results and a generalized version of agony \cite{Gupte2011} is used to uncover hierarchies. The approach is used to categorize teams into a few discrete levels of playing quality.
General match-up modeling is addressed by the \textit{blade-chest} model \cite{Chen2016a}. Each player is represented by two $d$-dimensional vectors, the \textit{blade} and \textit{chest} vectors. Team $a$ won if its blade is closer to team $b$'s chest than vice versa. Intransitivity is explicitly modeled by using both blade and chest vectors, something that cannot be accounted for by approaches that associate a single scalar value with each team \cite{Bradley1952}. 

\section{Soccer Team Vectors}
\label{sec:main}
In this section we present \textit{STEVE - Soccer Team Vectors}. We first give an overview of the underlying idea and the goal of this work. Afterwards we discuss the problem definition and introduce an algorithm to learn useful latent representations for soccer teams.

\subsection{Overview}
\label{sec:overview}
\textit{STEVE} aims to learn meaningful representations for soccer teams where representations come in the form of low dimensional vectors. If two teams are similar, their representations should be close in vector space while dissimilar teams should exhibit a large distance. Furthermore, these learned latent representations can be used as feature vectors for various machine learning tasks like clustering, classification and regression.
Due to the fact that there is no clear definition of similarity for soccer teams, we base our approach on the following four assumptions:
\begin{enumerate}
    \item The similarity between two teams can be determined by accounting for the matches they played in the past. 
    \item Frequent draws between two teams indicate that they are of approximately equal strength. Hence, both teams are similar.
    \item Two teams are similar if they often win against the same opponents.
    \item More recent matches have a higher influence on the similarity than those a long time ago.
\end{enumerate}
Since data acquisition is expensive and time-consuming, especially in the field of sports analytics, \textit{STEVE} is designed to learn from minimal information. More precisely, we only use data about which teams played against each other, during which season a match took place and whether the home team won, lost or the match resulted in a draw. Note that the assumptions mentioned above do not require any further information and are therefore well suited for this setting.

\subsection{Problem Definition}
To simplify definitions, let $M=\{1,2,\dots,m\}$, we assume that each of the $m$ soccer teams we want to learn a representation for is associated with an identification number $i\in M$. Further, let $\Phi \in \mathbb{R}^{m \times \delta}$, where each row $\Phi_i$ represents team $i$'s $\delta$ dimensional latent representation. The goal of this work is to find $\Phi$ in such a way, that $dist(\Phi_i, \Phi_j)$ is small for similar teams $i$ and $j$  and $dist(\Phi_i, \Phi_{k})$ is large for dissimilar teams $i$ and $k$. $dist(\cdot, \cdot)$ is some distance metric and similarity between teams is determined according to the assumptions made in Section \ref{sec:overview}. 
To solve this task, data is given in the following form: 
$\mathcal{D}=\{(a, b, s, d) \in M \times M \times \{1,\dots,x_{max}\} \times \{0,1\}\}$.
The quadruple $(a, b, s, d)$ represents a single match between teams $a$ and $b$, $s$ is an integer indicating during which of the $x_{max}$ seasons the match took place and $d$ is a flag set to $1$ if the match resulted in a draw and $0$ otherwise.
If $d=0$, the quadruple is arranged such that team $a$ won against team $b$.

\subsection{Algorithm}
According to the first assumption, we can loop over the dataset $\mathcal{D}$ while adjusting $\Phi$. If $d=1$, we minimize the distance between $\Phi_a$ and $\Phi_b$, thereby accounting for the second assumption.
The third assumption addresses a higher order relationship, where teams that often win against the same teams should be similar.
We introduce a second matrix $\Psi \in \mathbb{R}^{m \times \delta}$ and call each row $\Psi_i$ team $i$'s loser representation. Further, we call $\Phi_i$ the winner representation of team $i$. Both matrices $\Phi$ and $\Psi$ are initialized according to a normal distribution with zero mean and unit variance.
When team $a$ wins against team $b$ we minimize the distance between $\Phi_a$ and $\Psi_b$, bringing $b$'s loser representation and $a$'s winner representation closer together. That is, the loser representations of all teams $a$ often wins against, will be in close proximity to team $a$'s winner representation. Consequently, if other teams also often win against these teams, their winner representations must be close in order to minimize the distance to the loser representations. Parameters $\Phi$ and $\Psi$ are estimated using stochastic gradient descent where the objective we aim to minimize is given as follows:
$$
\argmin_{\Phi, \Psi}  \sum_{(a,b,s,d) \in \mathcal{D}} d * dist(\Phi_a, \Phi_b) + (1-d) * dist(\Phi_a, \Psi_b)
$$
We minimize the distance between $\Phi_a$ and $\Phi_b$ directly when both teams draw ($d=1$). Otherwise ($d=0$) we minimize the distance between $\Phi_a$ (winner representation) and $\Psi_b$ (loser representation).
With the squared euclidean distance as the distance metric, the expression can be rewritten as illustrated below.
$$
\argmin_{\Phi, \Psi}  \sum_{(a,b,s,d) \in \mathcal{D}}  d * \|\Phi_a - \Phi_b\|^2 + (1-d) * \|\Phi_a - \Psi_b\|^2
$$
In its current form, matches played in long past seasons contribute as much to the loss as more recent matches. We alleviate this problem by down-weighting matches from older seasons using the linear weighting scheme  $\frac{s}{x_{xmax}}$, thereby completing the formulation of the objective:
$$
\argmin_{\Phi, \Psi}  \sum_{(a,b,s,d) \in \mathcal{D}} \frac{s}{x_{max}} \Big[ d * \|\Phi_a - \Phi_b\|^2 + (1-d) * \|\Phi_a - \Psi_b\|^2 \Big]
$$
This approach has the advantage of no complex statistics having to be gathered. All our assumptions are captured in the teams's representations. We describe the algorithm in more detail in Algorithm \ref{algo:steve}. Note that here gradients are computed after observing a single data point and the regularization term is omitted. This is done for illustration purposes only. In our implementation, we train the algorithm in a batch-wise fashion. For
In lines 9, 12 and 15 the representations are normalized as we have found this to speed up training. It also helps to keep distances within a meaningful range.
\begin{algorithm}
\caption{STEVE($\mathcal{D}$, m, $\delta$, $\alpha$, $x_{max}$,e)}
\label{algo:steve}
\begin{algorithmic}[1]
    \State $\Phi \sim \mathcal{N}(0,1)^{m \times \delta}$ \Comment{Initialize $\Phi$}
     \State $\Psi \sim \mathcal{N}(0,1)^{m \times \delta}$ \Comment{Initialize $\Psi$}
    \For{$i$ in $\{1, \dots, e\}$}
        \State $\mathcal{D} = $ shuffle($\mathcal{D}$) \Comment{Shuffle dataset}
        \For {each $(a,b,s,d)$ in $\mathcal{D}$}
            \State $L(\Phi, \Psi) = \frac{s}{x_{xmax}} \Big[ d * \|\Phi_a - \Phi_b\|^2 + (1-d) * \|\Phi_a - \Psi_b\|^2 \Big]$ \Comment{Compute loss}
            \If {$d = 0$} \Comment{$a$ won the match}
                \State $\Psi_b = \Psi_b - \alpha * \frac{\partial L}{\partial \Psi_b}$ \Comment{Gradient descent on $b$'s loser representation}
                \State $\Psi_b = \Psi_b / \|\Psi_b\|_2$ \Comment{Normalize $b$'s loser representation}
            \Else \Comment{Match is a draw}
                \State $\Phi_b = \Phi_a - \alpha * \frac{\partial L}{\partial \Phi_b}$ \Comment{Gradient descent on $b$'s winner representation}
                \State $\Phi_b = \Phi_b / \|\Phi_b\|_2$ \Comment{Normalize $b$'s winner representation}
            \EndIf
            \State $\Phi_a = \Phi_a - \alpha * \frac{\partial L}{\partial \Phi_a}$ \Comment{Gradient descent on $a$'s winner representation}
            \State $\Phi_a = \Phi_a / \|\Phi_a\|_2$ \Comment{Normalize $a$'s winner representation}
    	\EndFor
	\EndFor
	\State \Return $\Phi$, $\Psi$
\end{algorithmic}
\end{algorithm}

\section{Experiments}
\label{sec:experiments}
In this section, we provide an overview of the dataset. We also conduct various experiments to investigate the expressiveness and efficacy of our approach. 

\subsection{Dataset and Experimental Setup}
The dataset consists of all the matches from the Bundesliga (Germany), Premier League (Great Britain), Serie A (Italy), La Liga (Spain), Eredivisie (Netherlands), League 1 (France), Süper Lig (Turkey), Pro League (Belgium), Liga NOS (Portugal), Europa League and the Champions League played from 2010 until 2019.
A total of 29529 matches between 378 different teams was carried out where approximately 25\% ended in a draw.
Unless stated otherwise, for all experiments we set $\delta=16$ and batch size $=128$. We use Adam~\cite{Kingma2014} with a learning rate $\alpha=0.0001$ for parameter estimation and train for $e=40$ epochs. Additionally, we add a small $L_2$ weight penalty of $10^{-6}$.

\subsection{Similarity Search} 
We select five European top teams and run $\textit{STEVE}$ on all the matches from season 2010 until 2019 in the corresponding league. Since we are dealing with small datasets, we set $\delta=10$ and batch size $=32$. For each team, we note the five most similar teams (smallest distance) in Table~\ref{tab:similarities}. Note that we use the distance between the corresponding winner representations. As expected, we clearly observe that top teams are similar to other top teams. For example, the team most similar to Barcelona is Real Madrid. Both teams often compete for supremacy in \textit{La Liga}. In general, we observe that similarities in Table \ref{tab:similarities} roughly reflect the average placement in the respective league. 

\begin{table}[h]

\caption{Five most similar teams for five European top teams.}\label{tab:similarities}
    \centering
    \setlength{\tabcolsep}{0.5em} 
    {\renewcommand{\arraystretch}{1.2}
    \begin{tabular}{|l|l|l|l|l|}
    \hline
    \multicolumn{5}{|c|}{\textbf{Top soccer team per league chosen for similarity search}}                                            \\ \hline
    Bayern M\"unchen & Barcelona       & Paris SG & Manchester U. & Juve. Turin        \\ \hline
    \multicolumn{5}{|c|}{\textbf{Five most similar teams chosen by \textit{STEVE}}}                                            \\ \hline
    RB Leipzig      & Real Madrid     & Lyon              & Liverpool         & SSC Napoli            \\ \hline
    Dortmund        & Valencia        & Marseille         & Manchester C.   & AS Roma               \\ \hline
    M\"onchengladbach & Atletico Madrid & Monaco            & Chelsea           & AC Milan              \\ \hline
    Leverkusen      & Sevilla         & St Etienne        & Tottenham         & Inter. Milano \\ \hline
    Hoffenheim      & Villarreal      & Lille             & Arsenal           & SS Lazio              \\ \hline
    \end{tabular}
    }

\end{table}

\subsection{Ranking Soccer Teams}
\begin{figure}[ht]

\centering
     \caption{Team rankings generated by \textit{STEVE}. Each row\textsuperscript{1,2} depicts one ranked list from the strongest (left) to the weakest team (right). Numbers represent a team's relative strength -  the number hypothetical matches won.}
     \label{fig:rankings}
     \includegraphics[width=1.0\textwidth]{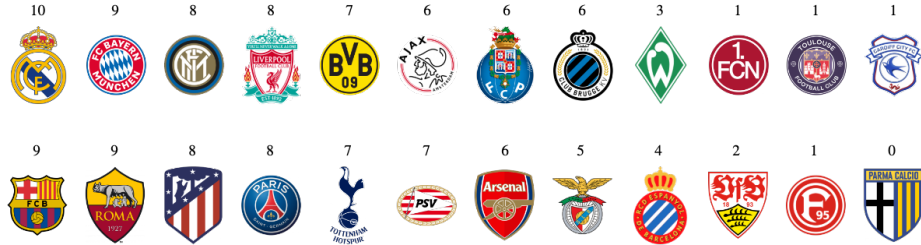}
     \scriptsize\textsuperscript{1} Real Madrid, FC Bayern Munich, Inter Milano, Liverpool FC, Borussia Dortmund, Ajax Amsterdam, FC Porto, Club Brugge KV, Werder Bremen, 1.FC Nuremberg, FC Toulouse, Cardiff City\\
     \scriptsize\textsuperscript{2} FC Barcelona, AS Roma, Atlético Madrid, Paris SG, Tottenham, PSV Eindhoven, Arsenal London, SL Benfica, Espanyol Barcelona, VFB Stuttgart, Fortuna Dusseldorf, Parma Calcio
\end{figure}

To retrieve a ranked list of soccer teams, one could simply use a league table. However, the list will only reflect the team's constitution accumulated over a single season. The ranking will not take past successes into account. One might alleviate this problem by averaging the league table over multiple seasons. Nevertheless, another problem arises: the list will only consist of teams from a single league. Combining league tables from different countries and competitions to obtain a more diverse ranking is considerably less straightforward. It gets even more complicated when we wish to rank a list of self chosen teams, possibly from many different countries. 
\textit{STEVE} provides a simple yet effective way to generate rankings for the use case mentioned above. Given a list of teams, we simulate a tournament where each team plays against all other teams. The list is then sorted according to the number of victories. To compute the outcome of a single match (victory or defeat) between team $a$ and $b$, let  $\alpha = \|\Phi_a - \Psi_b\|^2$ and $\beta = \|\Phi_b - \Psi_a\|^2$. If $\alpha < \beta$, then team $b$'s loser representation is closer to team $a$'s winner representation than team $a$'s loser representation is to team $b$'s winner representation. Thus, team $a$ is stronger than team $b$ and we increase team $a$'s victory counter. The same line of reasoning is applied to the case where $\alpha > \beta$.\\
In Figure~\ref{fig:rankings} we generated two rankings using the aforementioned approach. Each list consists of twelve teams from different European countries of different strengths. Our approach produces reasonable rankings: Highly successful international top teams like \textit{Real Madrid, FC Bayern Munich, FC Barcelona}, and \textit{AS Roma} are placed at the top of the list while mediocre teams like \textit{Espanyol Barcelona} and \textit{Werder Bremen} are placed further back in the list. The least successful teams like \textit{FC Toulouse, Cardiff City, Fortuna D\"usseldorf} and \textit{Parma Calcio} occupy the tail of the list.
\textit{STEVE} can be seen as an alternative to previous soccer team ranking methods \cite{Hvattum2010,Leitner2010} based on the ELO rating \cite{Elo1978}.

\vspace{-3mm}
\subsection{Team Market Value Estimation}
The goal of this work is to learn representations that are well suited for various downstream machine learning tasks. We validate this property by evaluating \textit{STEVE} with respect to regression and classification performance. We argue that a meaningful representation should carry enough information to reliably predict a team's market value.
Therefore, both tasks involve predicting the value of a team given its representation. 
We obtained current market values for all teams in the dataset from season 2018/2019. 
A team's market value is determined by the sum of the market values of all its players.
On average, a team is worth \euro 183.7 million with a standard deviation of \euro241.2M. The least valuable team is \textit{BV De Graafschap} (\euro10.15M) and the most valuable team is \textit{FC Barcelona} (\euro1180M). The first, second and third quartiles are \euro25M, \euro93.7M and \euro232.5M, respectively.
For regression and classification, we use the following team representations as an input to a multi layer perceptron (MLP) with two hidden layers. The first hidden layer has a size of $50$, the second one $20$. Apart from changing hidden layer sizes, we use default parameters provided by \cite{scikit-learn} for all further analyses.
\begin{itemize}
    \item \textbf{STEVE} 
    Representations are computed using \textit{STEVE} with $\delta \in \{8, 16, 32\}$.
    A team's winner and loser representation is concatenated to form its team vector.
    The resulting feature vectors are of size 16, 32, 64.
    \item \textbf{Season-stats} 
    We extract count based features for each team in the dataset to mimic traditional feature extraction.
    For season 2018/2019 we collect the following statistics: number of victories, draws, defeats as well as goals scored and goals conceded. Each feature is computed for matches in the Champions League, Euro League and the respective national league. Additionally, we use goals per match, goals per national and international match.
    This results in a $18$ dimensional feature vector (representation) for each team.
    \item \textbf{Season-stats (CAT-x)}
    Season-stats for the last $x$ seasons are concatenated. 
    The resulting feature vectors are of size $x*18$.
    \item \textbf{Season-stats (SUM-x)}
    Season-stats for the last $x$ seasons are summed together.
    The resulting feature vectors are of size $18$.
\end{itemize}

\noindent Comparability between the different representations mentioned above is ensured due to the fact that none of them requires information absent in the dataset.\\
\textit{Season-Stats} has many features that are intuitively well suited for team value estimation. For example, a large proportion of teams that participate in international competitions are more valuable than those who don't. Statistics about goals and match results are helpful for assessing a team's strength which is in turn correlated to the team's market value.\\

\noindent \textbf{Regression}
Team value estimation naturally lends itself to be cast as a regression problem. During training we standardize team values (targets) and \textit{Season-Stats} features by subtracting the mean and dividing by the standard deviation. Evaluation is carried out using 5-fold cross-validation and results are reported in Table~\ref{tab:regression}.\\

\noindent \textbf{Classification} 
By grouping team values into bins, we frame the task as classification problem. Teams are assigned classes according to the quartile their value lies in. Consequently, each team is associated with one of four classes. We apply the same standardization procedure as in the case of regression and use 5-fold cross-validation. Results are reported in Table~\ref{tab:classification}.\\

\begin{table}[ht]
\caption{Results for the regression task of team value estimation. To quantify the quality of prediction, we use root mean squared error (RMSE), mean absolute error (MAE) and the mean median absolute error (MMAE), all reported in million \euro.}\label{tab:regression}
\centering
\setlength{\tabcolsep}{0.75em} 
{\renewcommand{\arraystretch}{1.25}
\begin{tabular}{l|l|l|l|l|}
\cline{2-4}
                                           & RMSE                & MAE                & MMAE            \\ \hline
\multicolumn{1}{|l|}{STEVE-16}  & 142.12 $\pm$ 75.22 & 88.37 $\pm$ 25.69  &52.01 $\pm$ 13.42 \\ \hline
\multicolumn{1}{|l|}{STEVE-32}  & 131.51 $\pm$ 40.15  & 83.20 $\pm$ 24.51  & 46.87 $\pm$ 21.89\\ \hline
\multicolumn{1}{|l|}{STEVE-64}  & \textbf{111.27} $\pm$ 48.58  & \textbf{67.14} $\pm$ 30.51  & \textbf{32.80} $\pm$ 10.42 \\ \hline
\multicolumn{1}{|l|}{Season-Stats}         & 173.75 $\pm$ 119.35 & 110.32 $\pm$ 63.61 & 69.96 $\pm$ 15.43 \\ \hline
\multicolumn{1}{|l|}{Season-Stats (CAT-3)} & 200.77 $\pm$ 157.55 & 138.15 $\pm$ 87.06 & 86.98 $\pm$ 39.83 \\ \hline
\multicolumn{1}{|l|}{Season-Stats (CAT-5)} & 172.05 $\pm$ 70.83  & 119.81 $\pm$ 43.18 & 80.74 $\pm$ 18.96 \\ \hline
\multicolumn{1}{|l|}{Season-Stats (CAT-9)} & 151.09 $\pm$ 80.37  & 105.98 $\pm$ 41.96 & 68.82 $\pm$ 23.15 \\ \hline
\multicolumn{1}{|l|}{Season-Stats (SUM-3)} & 158.44 $\pm$ 108.50 & 105.65 $\pm$ 53.95 & 69.16 $\pm$ 11.39 \\ \hline
\multicolumn{1}{|l|}{Season-Stats (SUM-5)} & 154.71 $\pm$ 115.76 & 104.04 $\pm$ 59.34 & 69.81 $\pm$ 15.69  \\ \hline
\multicolumn{1}{|l|}{Season-Stats (SUM-9)} & 158.33 $\pm$ 120.90 & 106.67 $\pm$ 62.61 & 67.74 $\pm$ 17.75  \\ \hline
\end{tabular}
}

\end{table}

\begin{table}[ht!]
\caption{Results for the classification task of team value estimation, measured with micro $F_1$ score and macro $F_1$ score.}\label{tab:classification}
\centering
\setlength{\tabcolsep}{0.75em} 
{\renewcommand{\arraystretch}{1.25}
\begin{tabular}{l|l|l|}
\cline{2-3}
                                           & Micro $F_1$     & Macro $F_1$     \\ \hline
\multicolumn{1}{|l|}{STEVE-16}  & 0.67 $\pm$ 0.10 & 0.64 $\pm$ 0.10 \\ \hline
\multicolumn{1}{|l|}{STEVE-32}  & \textbf{0.74} $\pm$ 0.11 & 0.71 $\pm$ 0.14 \\ \hline
\multicolumn{1}{|l|}{STEVE-64}  & \textbf{0.74} $\pm$ 0.10 & \textbf{0.72} $\pm$ 0.09 \\ \hline
\multicolumn{1}{|l|}{Season-Stats}         & 0.52 $\pm$ 0.14 & 0.45 $\pm$ 0.19 \\ \hline
\multicolumn{1}{|l|}{Season-Stats (CAT-3)} & 0.50 $\pm$ 0.12 & 0.44 $\pm$ 0.15 \\ \hline
\multicolumn{1}{|l|}{Season-Stats (CAT-5)} & 0.55 $\pm$ 0.14 & 0.51 $\pm$ 0.16 \\ \hline
\multicolumn{1}{|l|}{Season-Stats (CAT-9)} & 0.60 $\pm$ 0.13 & 0.56 $\pm$ 0.11 \\ \hline
\multicolumn{1}{|l|}{Season-Stats (SUM-3)} & 0.49 $\pm$ 0.09 & 0.40 $\pm$ 0.10 \\ \hline
\multicolumn{1}{|l|}{Season-Stats (SUM-5)} & 0.48 $\pm$ 0.08 & 0.39 $\pm$ 0.07 \\ \hline
\multicolumn{1}{|l|}{Season-Stats (SUM-9)} & 0.47 $\pm$ 0.09 & 0.37 $\pm$ 0.15 \\ \hline
\end{tabular}
}

\end{table}

\noindent \textbf{Results}
\textit{STEVE} clearly outperforms the other representations both in terms of regression and classification performance. While $\delta = 64$ generally yields the best results, even $\delta=16$ produces superior results compared to \textit{Season-Stats}. In terms of regression performance, we observe that \textit{Season-Stats} is most competitive when using information from multiple seasons (CAT-x and SUM-x). All forms of representation manage to estimate the general tendency of a team's market value but \textit{STEVE's} predictions are far more precise.
Similar conclusions can be drawn when inspecting classification performance. The best competing representation is \textit{Season-Stats (CAT-9)} which is $162$ dimensional, $92$ dimensions more than \textit{STEVE ($\delta=64$)}.
Still, \textit{STEVE ($\delta=64$)} provides $\approx$ $20\%$ better performance than \textit{Season-Stats (CAT-9)}.
It can therefore be concluded that \textit{STEVE} is able to compress information needed for the task and succeeds to provide high efficacy representations.

\section{Conclusion}
\label{sec:conculsion}
In this work we introduced \textit{STEVE}, a simple yet effective way to compute meaningful representations for soccer teams. 
We provided qualitative analysis using soccer team vectors for team ranking and similarity search.
Quantitative analysis was carried out by investigating the usefulness of the approach by estimating the market values of  soccer teams. In both cases, \textit{STEVE} succeeds to provide meaningful and effective representations.
Future work might investigate further upon different weighting schemes for the season during which a match took place.
For example instead one can use the exponential distribution to weigh down past seasons.
Moreover, including the number of goals scored during a match and accounting for the home advantage might help to capture more subtleties. 
\bibliography{bibliography.bib}
\bibliographystyle{splncs04}
\end{document}